\documentclass[letterpaper, 10 pt, conference]{ieeeconf}  

\IEEEoverridecommandlockouts                              

\usepackage{graphics} 
\usepackage{epsfig} 
\usepackage{times} 
\usepackage{amsmath} 
\usepackage{amssymb}  
\usepackage{hyperref}

\usepackage{multirow}
\usepackage{booktabs}   
\title{\LARGE \bf
Differentiable Motion Manifold Primitives for Reactive Motion Generation under Kinodynamic Constraints
}

\author{Yonghyeon Lee
\thanks{Y. Lee is with the Department of Mechanical Engineering, Massachusetts Institute of Technology, Cambridge, MA 02139 USA (e-mail: yhl@mit.edu).}
}

\begin{document}

\maketitle
\thispagestyle{empty}
\pagestyle{empty}

\begin{abstract}
Real-time motion generation -- which is essential for achieving reactive and adaptive behavior -- under kinodynamic constraints for high-dimensional systems is a crucial yet challenging problem.  
We address this with a two-step approach: \emph{offline} learning of a lower-dimensional trajectory manifold of task‑relevant, constraint‑satisfying trajectories, followed by rapid \emph{online} search within this manifold.  
Extending the discrete‑time Motion Manifold Primitives (MMP) framework, we propose \emph{Differentiable Motion Manifold Primitives (DMMP)}, a novel neural network architecture that encodes and generates continuous‑time, differentiable trajectories, trained using data collected offline through trajectory optimizations, with a strategy that ensures constraint satisfaction -- absent in existing methods.
Experiments on dynamic throwing with a 7‑DoF robot arm demonstrate that DMMP outperforms prior methods in planning speed, task success, and constraint satisfaction.
Project page: \href{https://diffmmp.github.io/}{https://diffmmp.github.io/}.
\end{abstract}

\section{Introduction}
\begin{figure}[!t]
    \centering
    \includegraphics[width=\linewidth]{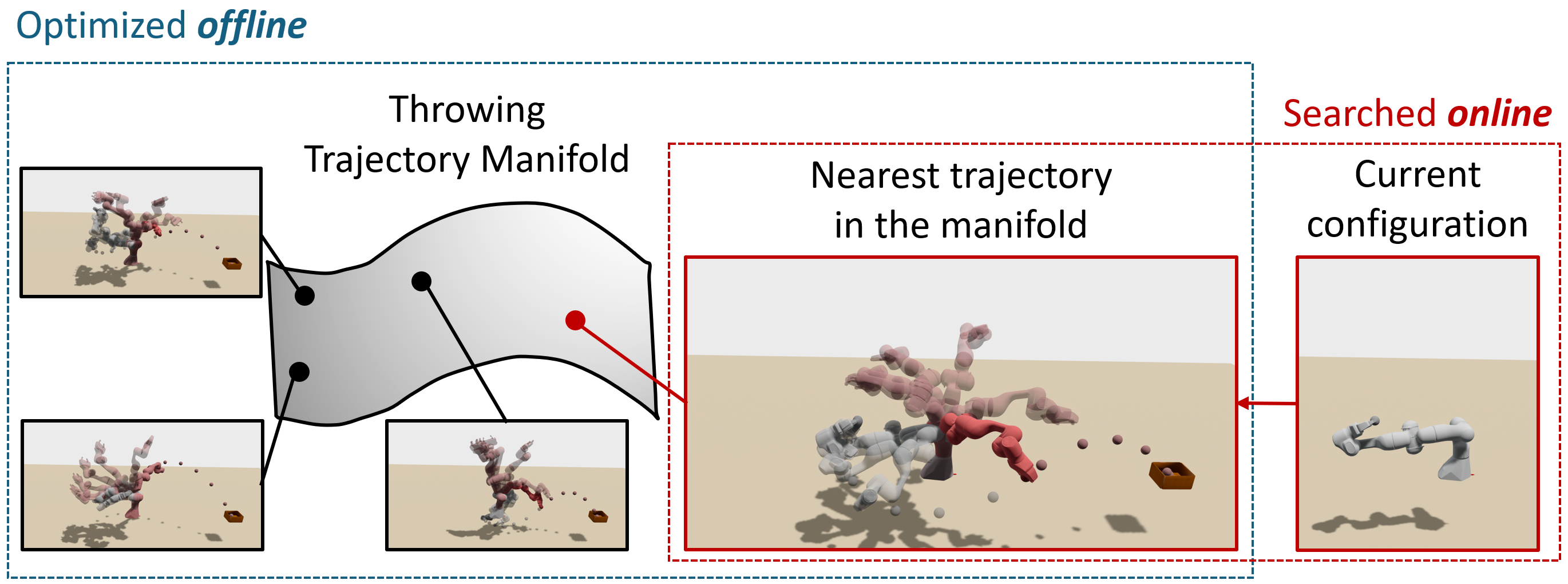}
    \caption{Throwing trajectory manifold -- which consists of task-completing trajectories that satisfy the kinodynamic constraints -- is optimized {\it offline}, and given a current configuration, the nearest throwing trajectory can be quickly searched {\it online} within the manifold. The moment of the throw is depicted in deep red, preceded by gray and followed by transparent red.}
    \label{fig:intro}
    \vspace{-10pt}
\end{figure}

Motion Manifold Primitives (MMP) is a generative movement primitive framework based on an autoencoder architecture~\cite{kingma2013auto,hinton2006reducing, bank2023autoencoders}. It encodes a diverse set of trajectories -- collected primarily from human demonstrations -- into a lower-dimensional manifold {\it offline}, thereby enabling {\it online} real-time motion generation and supporting reactive, adaptive behavior in dynamically changing environments~\cite{noseworthy2020task,lee2023equivariant,lee2024mmp++,lee2025motion}.
In this paper, our main objective is to extend this framework to motion generation under stringent kinodynamic constraints, where the generated motions are required to push against the limits of these constraints to successfully accomplish the task (e.g., see Fig.~\ref{fig:intro}).

Throughout, we assume that the objective function and constraints are given -- unlike previous approaches that rely on demonstration data -- so, in principle, a solution could be obtained by solving a trajectory optimization problem.   
However, we focus on problems whose solutions cannot be computed in real time -- specifically, not within sub‑second latency -- due to their complexity and the high dimensionality of the system, despite many advances in sampling, search, and optimization techniques~\cite{chen2020fuzzy,bonalli2019gusto,malyuta2022convex,sakcak2019sampling,yavari2019lazy,atreya2022state,qureshi2021constrained,qureshi2020motion}.

A natural approach is to adapt the MMP framework to our setting by collecting demonstration data through offline trajectory optimization.  
However, existing MMP methods do not explicitly incorporate kinodynamic constraints during training, which often results in trajectories that substantially violate these constraints.

To train an MMP model that satisfies the imposed constraints, we present two key contributions.  
The first is \emph{Differentiable Motion Manifold Primitives (DMMP)}, a novel neural network architecture designed to encode and generate a manifold of continuous‑time trajectories that are differentiable in time.  Unlike prior discrete‑time MMP formulations, this differentiability enables us to naturally enforce constraint satisfaction during training by incorporating the constraints directly into the loss function.

\begin{figure*}[!t]
    \centering
    \includegraphics[width=0.8\textwidth]{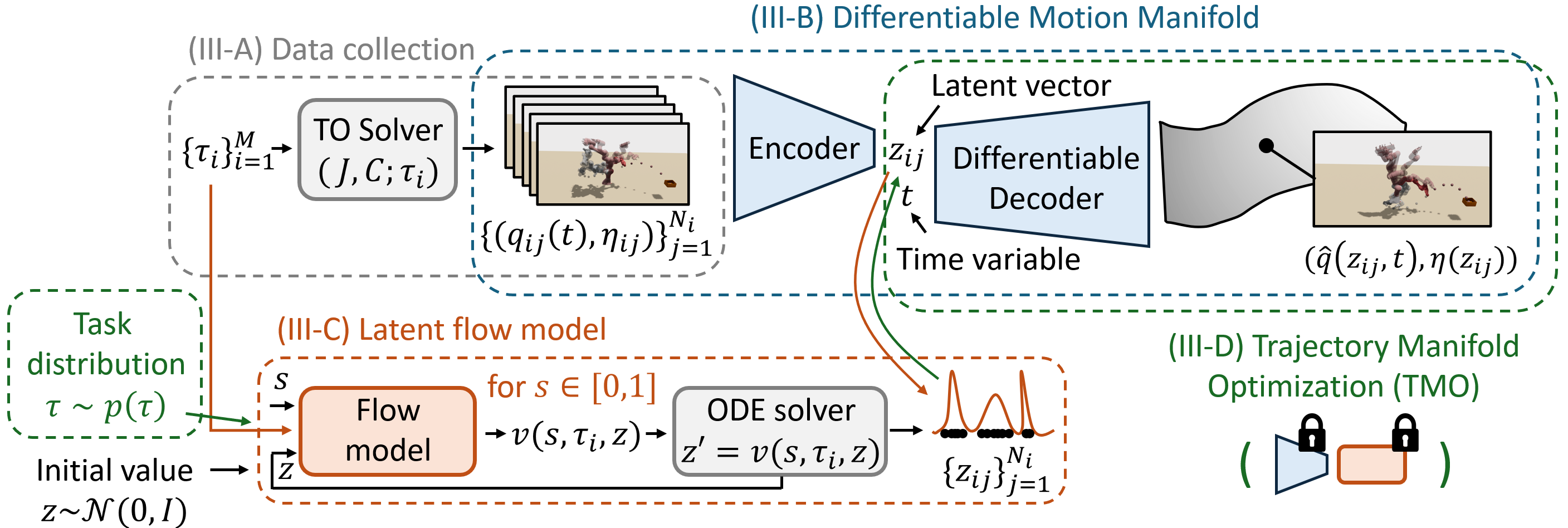}
    \caption{Illustration of the overall training pipeline for the trajectory manifold. First, given a finite set of task parameters, we collect multiple, diverse trajectories for each task parameter using any existing Trajectory Optimization (TO) method. Second, we fit a Differentiable Motion Manifold (DMM) consisting of an encoder and a differentiable decoder -- by differentiable, we mean differentiable in time. Third, we learn a flow-based, task-conditioned distribution in the latent space. Lastly, we fine-tune the trajectory manifold by optimizing the decoder -- where we freeze the pre-trained encoder and latent flow model -- to ensure that the generated trajectories achieve the tasks and comply with kinodynamic constraints for all task parameters randomly sampled from the task distribution.}
    \label{fig:summary}
\end{figure*}

The second contribution is a practical four‑step training strategy for DMMP (see Fig.~\ref{fig:summary}; details are provided in Section~\ref{sec:method}).  
First, we collect multiple trajectories for each task parameter by solving trajectory optimization problems with diverse random initializations and seeds.  
Second, we fit these trajectories to a differentiable motion manifold.  
Third, we train a task‑conditioned latent flow model, following~\cite{lee2025motion}.  
Finally, we freeze the latent flow and fine‑tune the manifold to ensure that the generated trajectories both accomplish the tasks and satisfy the kinodynamic constraints.

We present a case study on a dynamic throwing task using a 7‑DoF robot arm, which serves as a representative example of challenging kinodynamic motion‑planning problems.  
To throw beyond the nominal workspace limits, the robot must adopt a preparatory posture -- such as pulling its arm back -- and fully exploit its velocity limits.  
This requires optimizing the entire joint trajectory and the throwing time, rather than only the throwing phase, while simultaneously satisfying kinodynamic constraints, thereby making the problem highly complex.

We compare our DMMP with conventional trajectory optimization methods~\cite{powell1994direct,kraft1988software,kingma2014adam}, as well as existing MMPs~\cite{lee2023equivariant,lee2025motion}, evaluating task success rates, constraint satisfaction rates, and planning times. Our findings demonstrate that our method generates trajectories much more quickly than traditional trajectory optimization, with significantly higher success and constraint satisfaction rates compared to existing MMPs.

\section{Differentiable Motion Manifold Primitives}
\label{sec:method}

We begin with assumptions, notations, and problem definitions.  
Let the configuration be $q \in Q$ and the task parameter $\tau \in \mathcal{T}$ (e.g., the target box position for a throwing task).
For a fixed terminal time $T$, a smooth trajectory is denoted by $q(t)$ for $t \in [0,T]$, and an additional variable $\eta$ may be needed to fully specify a motion -- for example, in a throwing task, the release time $\eta \in (0,T)$ determines when the object is released at $q(\eta) \in Q$.  
The task-dependent objective function is $J(q(\cdot),\eta;\tau)$, and kinodynamic constraints such as self‑collisions, joint limits, and bounds on velocity, acceleration, jerk, and torque are expressed as $C(q,\dot q,\ddot q,\dddot q)\in\mathbb{R}^k \le 0$ with $C$ differentiable.  
We assume that, for each $\tau$, multiple globally or locally optimal motions $(q(t),\eta)$ exist.  
Our goal is to learn a model, given any $\tau \in \mathcal{T}$, generates multiple feasible motions $(q(t),\eta)$ that satisfy the constraints and optimize the objective.

Our framework consists of four steps. First, we collect multiple optimal trajectories by solving trajectory optimizations for each $\tau$ in section~\ref{sec:A}. Using these trajectories, we fit a differentiable motion manifold in section~\ref{sec:B}. 
In section~\ref{sec:C}, we train a task-conditioned motion distribution in the latent space. 
Lastly, in section~\ref{sec:D}, we fine-tune the motion manifold. The overall pipeline is visualized in Fig.~\ref{fig:summary}.

\subsection{Data Collection via Trajectory Optimizations}
\label{sec:A}
We collect multiple trajectories for each $\tau$ to enable subsequent manifold learning.  
Since sampling all $\tau \in \mathcal{T}$ is infeasible, we select a finite subset $\mathcal{T}_s=\{\tau_i\}_{i=1}^M$ that approximates $\mathcal{T}$.  
For each $\tau_i$, we solve
\begin{equation}
\min_{q(t),\eta} J(q(\cdot),\eta;\tau_i) \quad 
\text{s.t. } C(q,\dot q,\ddot q,\dddot q)\le0,\ \forall t\!\in[0,T].
\end{equation}
Specifically, we parameterize the trajectory as
\begin{equation}
\label{eq:parmetric_model}
q(t)=q_0+(q_T-q_0)(3-2s)s^2+s^2(s-1)^2\Phi(s)w,
\end{equation}
where $s=\tfrac{t}{T}$ and $q_0,q_T\in Q\subset\mathbb{R}^n$. 
The basis matrix is $\Phi(s)=[\phi_1(s), \ldots, \phi_{B}(s)] \in \mathbb{R}^{1\times B}$, where $\phi_i(s) = \exp(-B^2(s-\frac{i-1}{B-1})^2)$ and $B$ is the number of basis functions. The coefficient matrix $w\in\mathbb{R}^{B\times n}$. 
This parameterization satisfies the boundary conditions $q(0)=q_0$, $q(T)=q_T$, and $\dot q(0)=\dot q(T)=0$.  
We optimize over $(q_0,q_T,w)$ with diverse random seeds and initializations to collect diverse solutions $\{(q_{ij}(t),\eta_{ij})\}_{j=1}^{N_i}$.

\subsection{Learning Differentiable Motion Manifold}
\label{sec:B}
\begin{figure}[!t]
    \centering
    \includegraphics[width=0.9\linewidth]{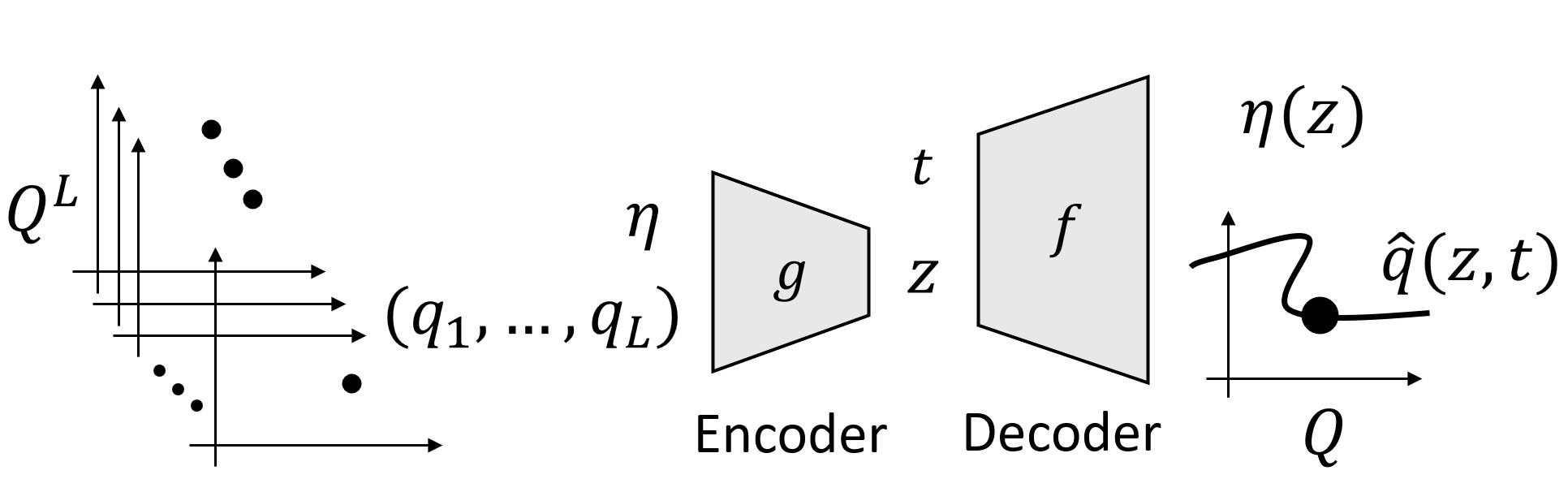}
    \caption{The encoder takes $(q_1,\ldots,q_L) \in Q^L$ and $\eta$ and outputs the latent value $z$; the decoder maps $z$ and time variable $t$ to a configuration $\hat{q}(z,t) \in Q$ and $\eta(z)$.}
    \label{fig:dmm}
\end{figure}

In this section, we propose a method to train a manifold of continuous-time, differentiable trajectories from the dataset $\{(q_{ij}(t),\eta_{ij})\}_{j=1}^{N_i}$.  
Following autoencoder-based manifold learning approaches~\cite{arvanitidis2017latent,lee2021neighborhood,yonghyeon2022regularized,lee2023explicit} and discrete-time MMPs~\cite{lee2023equivariant,lee2025motion}, we introduce an encoder $g$ and decoder $f$ with latent space $Z=\mathbb{R}^m$.  
The encoder maps a discretized trajectory $(q_1,\ldots,q_L)\in Q^L$ and $\eta$ to a latent variable $z\in Z$, i.e., $g((q_1,\ldots,q_L),\eta)=z$.  
Unlike discrete MMP decoders that take only $z$ and output a vectorized trajectory $(\hat{q}_1,\ldots,\hat{q}_L) \in Q^L$, our decoder takes $(z,t)$ and outputs $\hat q(z,t) \in Q$ and $\eta(z)$, with $\eta$ depending solely on $z$.  
Because $f(z,t)$ is differentiable in $t$, we refer to it as a \emph{differentiable decoder} (see Fig.~\ref{fig:dmm}).

The encoder and decoder are approximated with deep neural networks, denoted by $g_{\alpha}$ and $f_{\beta}=(\hat{q}_{\beta}, \eta_{\beta})$, respectively, with parameters $\alpha,\beta$, and trained to minimize the following loss function:
\begin{align}
    \label{eq:reconloss}
    {\cal L}_{{\rm recon}}(\alpha, \beta) := \frac{1}{M} \sum_{i} \frac{1}{N_i} \sum_{j} & \|\hat{q}_{\beta}(z_{ij}, t) - q_{ij}(t)\|_{c(t)}^2 \nonumber \\ & + \|\eta_\beta(z_{ij}) - \eta_{ij}\|^2, 
\end{align}
where $z_{ij} = g_{\alpha}((q_{ij}(t_1),\ldots, q_{ij}(t_L)), \eta_{ij})$, $t_l = \frac{l-1}{L-1}T$ for $l=1,\ldots, L$, and $\|\delta(t)\|_{c(t)}^2:=\int_{0}^T c(t) \delta^T(t) \delta(t) dt$ for some positive function $c(t)>0$.  
The positive function \( c(t) \) is introduced to fit more accurately on important parts along the time axis. For example, in the throwing task, the time around the throwing moment \( \eta \) should be given more weight.

Any neural network can be used for $f_\beta$.  
Inspired by deep operator learning~\cite{lu2019deeponet}, we adopt a linear basis function architecture for $\hat q_\beta$ that improves memory and computation efficiency:
\begin{equation}
\label{eq:lbfnn}
    \hat q_\beta(z,t) = \sum_{b=1}^{N_b} \psi_\beta^b(z)\,\theta_\beta^b(t),
\end{equation}
where $\psi_\beta^b: Z \to \mathbb{R}$ and $\theta_\beta^b:\mathbb{R}\to\mathbb{R}^n$ for $b=1,\ldots,N_b$ are neural networks.  
This structure avoids differentiating $\psi$ when computing time derivatives, reducing cost, and requires storing only $\psi(z)$ -- not the full network -- when inferring a trajectory for a fixed $z$, improving memory efficiency.

Given a sufficiently low-dimensional latent space $Z$ and the injective immersion conditions on $\hat{q}_{\beta}\colon z \mapsto \hat{q}_{\beta}(z, \cdot)$, we can interpret the set of trajectories $\{\hat{q}_{\beta}(z, \cdot) \mid z \in Z\}$ as an $m$-dimensional differentiable manifold of trajectories~\cite{arvanitidis2017latent,lee2023geometric,lee2022statistical,lee2024mmp++,yonghyeon2022regularized,heo2025isometric}.  
Although these conditions may not be strictly satisfied everywhere, they hold almost everywhere due to the smoothness properties of neural networks.  
Following established terminology~\cite{lee2023equivariant,lee2024mmp++,lee2025motion}, we adopt the term \emph{Differentiable Motion Manifold (DMM)} to denote this structure, as it captures the intrinsic low-dimensional manifold underlying the set of trajectories encoded by the model.

\subsection{Latent Flow Learning}
\label{sec:C}
This section describes how to fit a task‑conditioned density model $p(z|\tau)$ in the latent space of the learned differentiable motion manifold, adopting latent flow learning from MMFP~\cite{lee2025motion}.  
Let $z_{ij}$ be the latent encoding of $(q_{ij}(t),\eta_{ij})$ obtained from the trained encoder $g_{\alpha}$, giving a dataset of task–latent pairs $\big(\tau_i,\{z_{ij}\}_{j=1}^{N_i}\big)_{i=1}^M$.
Our goal is to train $p(z|\tau)$ so that, given a task $\tau$, we can sample $z$ and generate a motion via the decoder $f_\beta$, i.e., $z \mapsto f_\beta(z,\cdot)$.

Because $p(z|\tau)$ may be multimodal and highly nonconvex, we employ a flow‑based generative model~\cite{chen2018neural,lipman2022flow} for sufficient expressiveness.  
In latent space $Z$, we define a neural velocity field $v_\gamma(s,\tau,z)$ with parameters $\gamma$ and evolve it by
\begin{equation}
\label{eq:lsode}
    \frac{dz}{ds} = v_\gamma(s,\tau,z), \quad s \in [0,1],
\end{equation}
treating $p_\gamma(z|\tau)$ as the pushforward of a Gaussian prior $p_0(z)$.  
Sampling from $p_\gamma(z|\tau)$ involves drawing $z_0\sim p_0$ and integrating the ODE from $s=0$ to $s=1$.  
We train $v_\gamma$ using flow matching~\cite{lipman2022flow}, a simulation‑free method that is more efficient than maximum‑likelihood training.

Finally, by sampling $z \sim p_\gamma(z|\tau)$ and decoding with $f_\beta$, we generate motions for any given $\tau$.  
We refer to this framework as \emph{Differentiable Motion Manifold Flow Primitives (DMMFP)}.

\subsection{Trajectory Manifold Optimization}
\label{sec:D}
Trajectories generated by DMMFP are not guaranteed to satisfy kinodynamic constraints.  
Moreover, since the dataset covers only a finite set of task parameters \( \{\tau_i\}_{i=1}^M \), performance on unseen \( \tau \in \mathcal{T} \) is not guaranteed.  
To address this, we propose \emph{Trajectory Manifold Optimization (TMO)} to fine‑tune the manifold so that generated trajectories both achieve the task for all $\tau \in \mathcal{T}$ and satisfy the constraints.

Let $g_\alpha$, $f_\beta$, and $p_\gamma$ be the pre‑trained encoder, decoder, and latent flow.  
We fine‑tune only the decoder parameters $\beta$, keeping $\alpha$ and $\gamma$ fixed.  
Jointly tuning all parameters would require backpropagation through the sampled latent values and the ODE solver, making training computationally prohibitive, whereas adjusting $\beta$ alone is both efficient and sufficient to modify the manifold.

Define $U(S)$ as the uniform distribution over a set $S$.  
We introduce the task loss
\begin{align}
\label{eq:taskloss}
    {\cal L}_{\rm task}(\beta) := \mathbb{E}_{t, \tau, z}\Big[ &J(\hat{q}_\beta(z, \cdot), \eta_\beta(z); \tau) \nonumber \\ &+ W^T \big({\rm ReLU} (C(t, z, \beta))^2\big) \Big],
\end{align}
where $t \sim U([0,T])$, $\tau \sim U(\mathcal{T})$, and $z \sim p_\gamma(z|\tau)$.  
Here,
\begin{equation}
C(t,z,\beta) = C\!\big(\hat q_\beta, \tfrac{\partial}{\partial t}\hat q_\beta, \tfrac{\partial^2}{\partial t^2}\hat q_\beta, \tfrac{\partial^3}{\partial t^3}\hat q_\beta\big) \in \mathbb{R}^k,
\end{equation}
and $W \in \mathbb{R}^k$ is a positive weight vector.  
The first term in \eqref{eq:taskloss} enforces task success, while the second penalizes constraint violations.  
Sampling $\tau \sim U(\mathcal{T})$ improves generalization, as the model is no longer limited to the subset $\mathcal{T}_s$ previously used for training.

Finally, adding the reconstruction loss \eqref{eq:reconloss} with fixed $\alpha$, we minimize
\begin{equation}
{\cal L}(\beta) = w_{\rm recon}{\cal L}_{\rm recon}(\beta) + {\cal L}_{\rm task}(\beta),
\end{equation}
where $w_{\rm recon}>0$.  
We refer to this fine‑tuning process as \emph{Trajectory Manifold Optimization (TMO)}.

\section{Case Study: Dynamic Throwing \\ with a 7-DoF Robot Arm}

In this section, we demonstrate our method on a dynamic throwing task with a 7‑DoF Franka Panda arm, a challenging scenario that requires solving a kinodynamic optimization with nonlinear objectives and tight constraints.  
The solution depends heavily on initial conditions and is computationally expensive, making it ideal to showcase the benefits of our approach.

\begin{figure}[!t]
    \centering
    \includegraphics[width=0.9\linewidth]{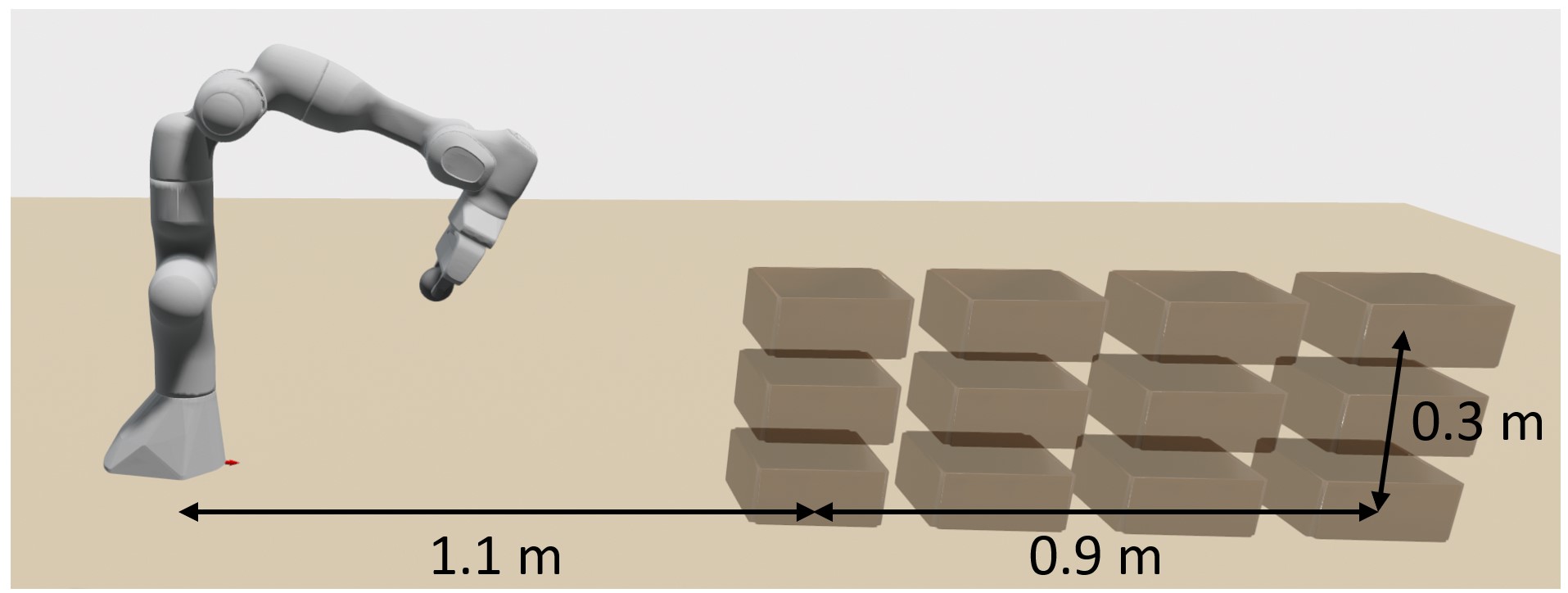}
    \caption{Task parameter space ${\cal T}$, a set of positions of the target box.}
    \label{fig:target_position}
\end{figure}

\textbf{Task setup.}  
As shown in Fig.~\ref{fig:target_position}, the task parameter is the target box position $\tau = (r\cos\theta, r\sin\theta, h)\in\mathbb{R}^3$ from the robot base.  
We exploit rotational symmetry by fixing $\theta=0$, since nonzero $\theta$ can be handled by rotating the first joint.  
Therefore, the task parameter space for training models is 
\[
{\cal T} = \{(r,0,h)\mid r\!\in\![1.1,2.0],\ h\!\in\![0.0,0.3]\},
\]
excluding close targets where throwing is unnecessary.  
The throwing time $\eta \in [0,T]$ with $T=5$ is an additional variable, and the release phase is $(q(\eta),\dot q(\eta))$.  
We assume the object is released instantaneously at time~$\eta$, with the velocity determined by the end‑effector at that instant.

\textbf{Objective.}  
Using free‑fall dynamics, the landing position is computed and the objective is defined as
\begin{equation}
J(q(\cdot),\eta;\tau) = J_{\rm task}(q(\cdot),\eta;\tau) + w_1 \int_0^T \|\dddot q(t)\|^2 dt,
\end{equation}
where $J_{\rm task}$ is the squared position error when the thrown object crosses the target box’s $z$‑level (height).

\textbf{Constraints.}  
We enforce limits on joint position, velocity, acceleration, and jerk; end‑effector velocity limits; torque limits derived from the dynamics; and self‑collision margins using capsule models.  
All constraints are expressed as $C(q,\dot q,\ddot q,\dddot q)\le0$, with added safety margins.  
We use the standard limits of the Franka Emika Panda, except that the end‑effector velocity limits are doubled to enable throws of up to 2\,m (in simulation).

\textbf{Implementation.}  
We weight trajectory errors by $c(t)=\exp\!\big(-4(t-\eta)^2\big)$ in (\ref{eq:reconloss}) to achieve higher accuracy near the throwing time.
In the basis model (\ref{eq:lbfnn}), $N_b=100$.  
All modules use fully connected networks with GELU activations (4-6 layers, 1024 nodes).  
The subset of task parameters for optimization is
\[
{\cal T}_s=\{(r,0,h)\mid r\!\in\!\{1.1,\dots,2.0\},\ h\!\in\!\{0.0,0.1,0.2,0.3\}\}.
\]
All kinematics, dynamics, and constraints are implemented in PyTorch. The latent ODE~(\ref{eq:lsode}) is integrated using the Euler method with step size $ds=0.1$, providing a good balance between accuracy and speed.

\subsection{Data Collection via Trajectory Optimizations} 
\label{sec:dcto}

We use the parametric trajectory model $q_{q_0,q_T,w}$ (\ref{eq:parmetric_model}) with $B=20$, so $w \in \mathbb{R}^{20 \times 7}$.  
The optimization variable $w$ is initialized to zero, and $\eta$ is set to $2$.  
To randomly initialize $q_0$ and $q_T$ within joint limits, we sample $v_0,v_T \in \mathbb{R}^7$ from a standard Gaussian, apply the sigmoid function, and scale them by the joint limits.  
Empirically, initialization in the vicinity of the origin leads to improved performance compared to uniform sampling, since extreme initial configurations can adversely affect convergence during optimization.

\begin{figure}[!t]
    \centering
    \includegraphics[width=\linewidth]{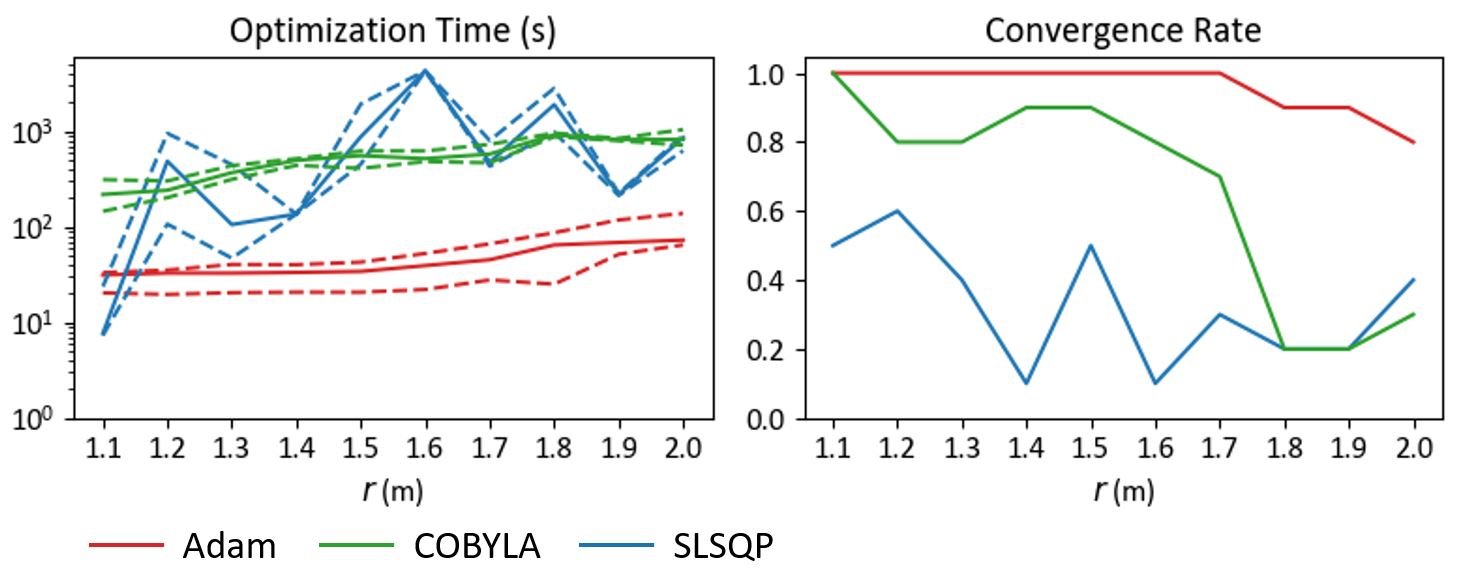}
    \caption{Quartiles of the optimization times and convergence rates as functions of \( r \in [1.1, 2.0] \), where the target box position is \(\tau = (r, 0, 0)\). The quartiles are computed using only the successful cases from 10 optimization attempts.}
    \label{fig:opt_graph}
\end{figure}

We first compare trajectory optimizers SLSQP~\cite{kraft1988software}, COBYLA~\cite{powell1994direct}, and Adam~\cite{kingma2014adam} by computation time and success rate.  
For Adam, constraints are added to the objective via a ReLU penalty.  
Optimization stops when constraints meet thresholds and position error $<0.01$, or after 10,000 iterations.  
Fig.~\ref{fig:opt_graph} shows that all solvers take over 10 s and are sensitive to initialization; Adam performs best but still fails beyond $r=1.7$, highlighting the problem’s complexity and the challenge of achieving sub‑second planning.

\begin{figure}[!t]
    \centering
    \includegraphics[width=\linewidth]{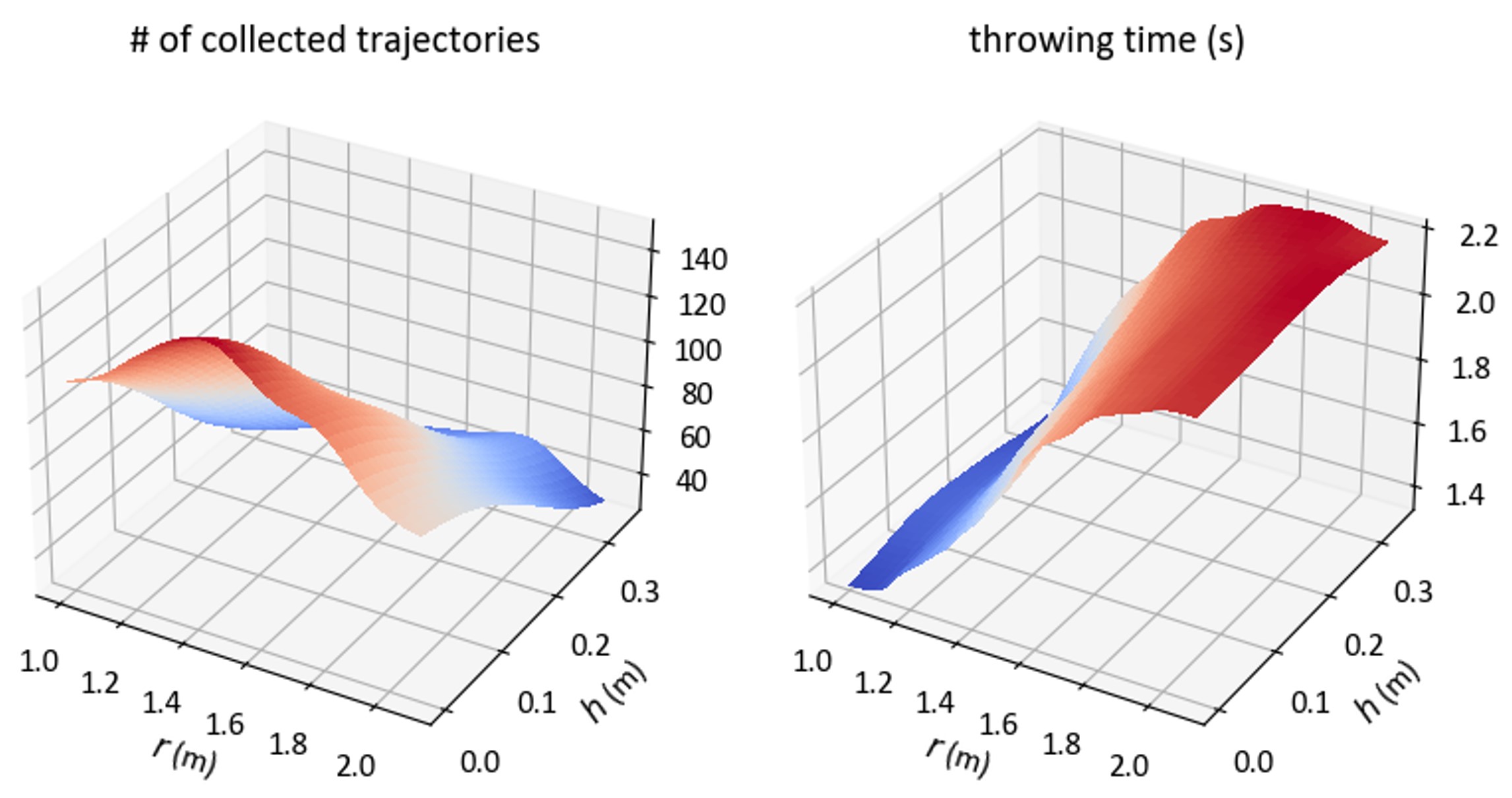}
    \caption{The number of collected trajectories and throwing times as functions of $r$ and $h$, where the target box position is $\tau = (r, 0, h)$.}
    \label{fig:datastat.}
\end{figure}

We collect trajectories by repeatedly solving the optimization with the Adam optimizer.  
For each $\tau \in {\cal T}_s$, 300 trajectories are optimized, yielding 12,000 candidates across 40 parameters.  
After excluding failure cases, 3,523 valid trajectories remain.  
Fig.~\ref{fig:datastat.} ({\it Left}) shows their distribution over task parameters, and Fig.~\ref{fig:datastat.} ({\it Right}) shows the mean throwing time $\eta$ as a function of the task parameters.  
As expected, fewer trajectories are obtained and $\eta$ increases as the target distance grows.

\subsection{Comparisons of Planning Performance}

\begin{table*}[!t]
\tiny
\centering    
\caption{Average success rates, position error, and various constraint satisfaction rates. Computations are performed using an AMD Milan 2.8 GHz 8-core processor for trajectory optimizations and AMD Ryzen 9 5900x 12-Core Processor and NVIDIA GeForce RTX 3090 for motion manifold primitives.}
\label{tab:main_table}
\begin{tabular}{lcccccccccccccccccccc}
\toprule
& \multicolumn{9}{c}{Seen Task Parameter} & \multicolumn{9}{c}{Unseen Task Parameter}  & \multirow{2}{*}{$\#$ Traj.} & \multirow{2}{*}{Time} \\ 
\cmidrule(lr){2-10} \cmidrule(lr){11-19}
&  SR & Error & JL & JVL & JAL & JJL & CVL & JTL & COL & SR & Error & JL & JVL & JAL & JJL & CVL & JTL & COL &  &\\ 
\midrule 
TO (Adam)~\cite{kingma2014adam} & 97 & 0.01 & 100 & 100 & 100 & 100 & 100 & 100 & 100 & 100 & 0.01 & 100 & 100 & 100 & 100 & 100 & 100 & 100 & 1 & 10 $\sim$ 100 s \\
TO (SLSQP)~\cite{kraft1988software} & 33 & 0.01 & 100 & 100 & 100 & 100 & 100 & 100 & 100 & 100 & 0.01 & 100 & 100 & 100 & 100  & 100 & 100 & 100 & 1 & 10 $\sim$ 3000 s \\
TO (COBYLA)~\cite{powell1994direct} & 66 & 0.01 & 100 & 100 & 100 & 100 & 100 & 100 & 100 & 100 & 0.01 & 100 & 100 & 100 & 100 & 100 & 100 & 100 & 1 & 100 $\sim$ 1000 s \\
MMP~\cite{lee2023equivariant} & 1.05 & 0.53 & 0.95 & 0.0 & 0.0 & 0.0 & 0.0 & 0.0 & 92.6 & 0.81 & 0.53 & 96.4 & 0.0 & 0.0 & 0.0 & 0.0 & 0.0 & 94.4 & 100 & 0.003s \\
MMFP~\cite{lee2025motion} & 77.4 & 0.05 & 97.7 & 0.0 & 0.0 & 0.0 & 0.0 & 0.0 & 99.0 & 15.0 & 0.15 & 0.0 & 0.0 & 0.0 & 0.0 & 0.0 & 0.0 & 99.3 & 100 & 0.011s\\
DMMFP & 17.5 & 0.18 & 99.2 & 72.5 & 25.9 & 100 & 89.5 & 64.7 & 98.6 & 4.96 & 0.26 & 99.4 & 79.5 & 32.7 & 100 & 93.1 & 87.4 & 98.4 & 100 & 0.012s\\
DMMFP + TMO & 95.8 & 0.01 & 100 & 93.0 & 99.9 & 100 & 99.9 & 100 & 100 & 94.1 & 0.02 & 100 & 80 & 98.2 & 100 & 100 & 100 & 100 & 100 & 0.012s \\ 
DMMFP + TMO + RS & 100 & 0.01 & 100 & 100 & 100 & 100 & 100 & 100 & 100 & 100 & 0.01 & 100 & 100 & 100 & 100 & 100 & 100 & 100 & 91 & 0.227s  \\ 
\bottomrule
\end{tabular}
\end{table*}

\begin{figure*}[!t]
    \centering
    \includegraphics[width=\textwidth]{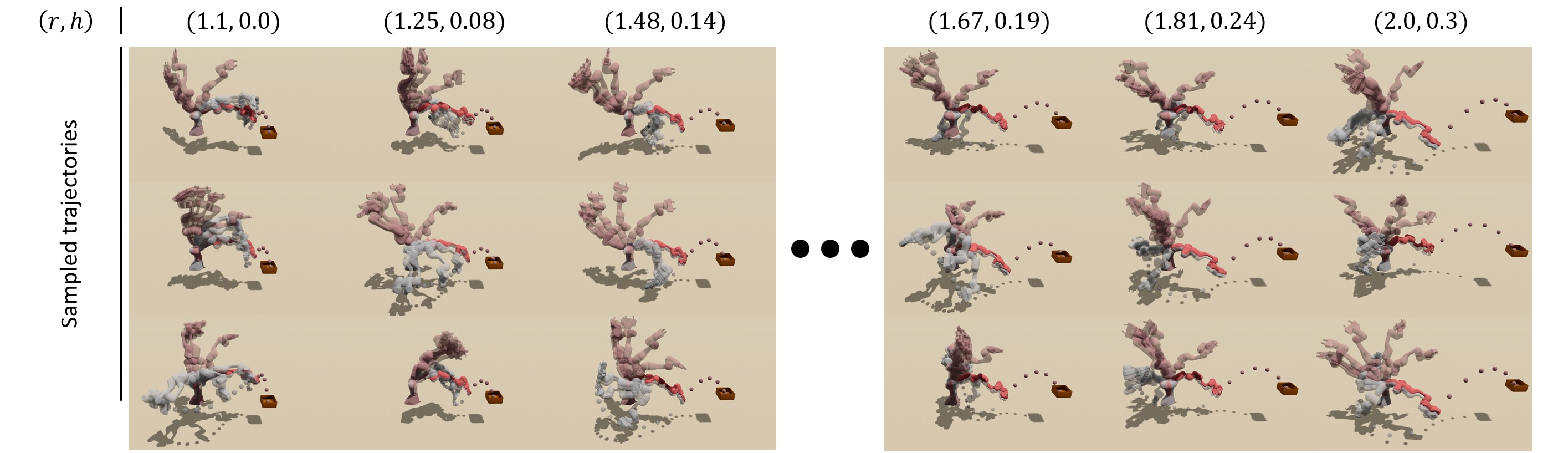}
    \caption{Examples of diverse generated trajectories by {\it Differentiable Motion Manifold Primitives (DMMFP) + Trajectory Manifold Optimization (TMO) + Rejection Sampling (RS)} for various task parameters $\tau=(r,0,h)$. The moment of the throw is depicted in deep red, preceded by gray and followed by transparent red.}
    \label{fig:results1}
\end{figure*}

\begin{figure*}[!t]
    \centering
    \includegraphics[width=0.8\textwidth]{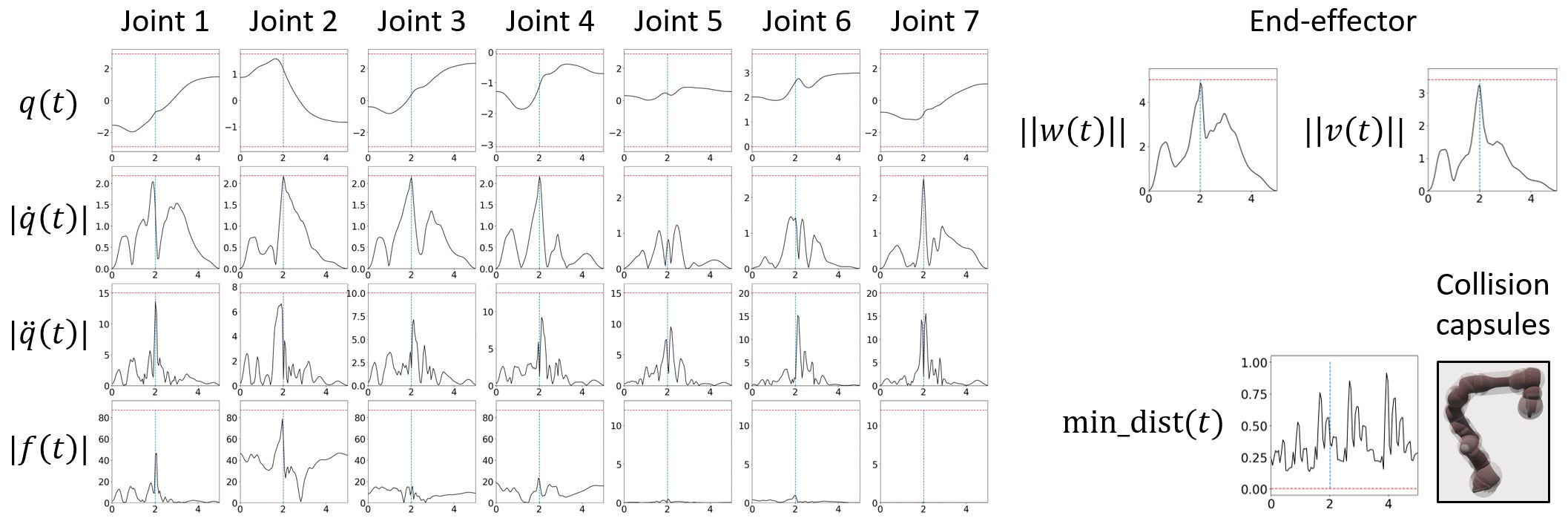}
    \caption{Trajectories of joint $q(t)$, joint velocity $\dot{q}(t)$, joint acceleration $\ddot{q}(t)$, joint torque $f(t)$, end-effector angular velocity $w(t)$, end-effector linear velocity $v(t)$, and minimum distance between links ${\rm min}\_{\rm dist}(t)$ for an example throwing motion generated by {\it DMMFP + TMO + RS} given the task parameter $\tau=(2.0, 0, 0.3)$. The limit values are visualized as red horizontal lines. The moment of throw is marked in blue vertical lines.}
    \label{fig:results2}
\end{figure*}

In this section, we compare {\it Trajectory Optimization (TO)} with the parametric curve models detailed in section~\ref{sec:dcto}, and motion manifold-based methods: {\it Motion Manifold Primitives (MMP)} with a Gaussian mixture prior~\cite{lee2023equivariant}, {\it Motion Manifold Flow Primitives (MMFP)}~\cite{lee2025motion}, and our {\it Differentiable Motion Manifold Flow Primitives (DMMFP)}, each of which is trained with a $32$-dimensional latent space. To train MMP, MMFP, and DMMFP, we use the same trajectory dataset collected via trajectory optimizations in section~\ref{sec:dcto}. We fine-tune the DMMFP using {\it Trajectory Manifold Optimization (TMO)} introduced in section~\ref{sec:D}, and denote it by ${\it DMMFP + TMO}$. 
Some generated trajectories by DMMFP + TMO may not meet the required constraints or task goals. Therefore, we use {\it Rejection Sampling (RS)} to exclude non-compliant samples, denoted by {\it DMMFP + TMO + RS}, although this step necessitates extra computation to verify the feasibility of the trajectories.

TABLE~\ref{tab:main_table} presents the averaged Success Rate (SR) -- a motion \((q(t), \eta)\) is deemed successful if the position error is less than 0.04 --, Position Error (Error) in meter scale, and Constraint Satisfaction Rates on Joint Limits (JL), Joint Velocity Limits (JVL), Joint Acceleration Limits (JAL), Joint Jerk Limits (JJL), Cartesian Velocity Limits (CVL), Joint Torque Limits (JTL), and Self-Collisions (COL).
A trajectory is considered to satisfy a constraint if it meets the requirement at every time \(t \in [0,T]\) along the trajectory. 
These metrics are computed using both seen task parameters $\tau \in {\cal T}_s$ and unseen test task parameters $\tau \in \{(r,0,h)| r \in \{1.15, 1.25, \ldots, 1.95\}, h \in \{0.05, 0.15, 0.25 \} \}$.
The number of trajectories generated simultaneously is reported as $\#$ Traj. in these experiments, along with the corresponding time consumed. 

From this table, we derive four key findings.  
First, trajectory optimization is significantly slower than manifold-based methods, highly sensitive to initialization, and can take over a minute as the target distance increases, making it impractical for online replanning (see Fig.~\ref{fig:opt_graph}).  
In contrast, motion‑manifold methods are much faster, as neural networks enable immediate sampling, even for 100 trajectories in parallel on a GPU.

Second, MMP, MMFP, and DMMFP all show low constraint‑satisfaction rates, indicating that fitting data alone is insufficient.  
DMMFP performs relatively better, benefiting from inductive bias that enforces some temporal smoothness.

Third, applying TMO markedly improves DMMFP’s performance.  
Residual failures are mainly due to Joint Velocity Limits (JVL), but using Rejection Sampling (RS) raises the success rate to 100\%, retaining about 91 of 100 samples.

Lastly, RS slightly increases runtime because constraint checking is required: our implementation takes about 0.2 s to verify 100 trajectories across 100 time points, involving forward kinematics, inverse dynamics, and 45 constraint checks for each $(q,\dot q,\ddot q,\dddot q)$ combination.  
All code is implemented in Python with PyTorch, measured on an AMD Ryzen 9 5900X and an NVIDIA RTX 3090.  
Although not fully optimized, further acceleration may be achieved through lower‑level implementations or by training a neural classifier to quickly predict trajectory feasibility.

Diverse throwing motions generated by {\it DMMFP + TMO + RS} are visualized in Fig.~\ref{fig:results1}.  
If only a single trajectory were available, the robot would often need to make inefficient adjustments, especially when its current configuration is far from that trajectory.  
In contrast, the diversity provided by our approach enables selecting a trajectory that better matches the current configuration.

An example motion for $\tau = (2.0,0,0.3)$ is shown in Fig.~\ref{fig:results2}, illustrating joint angles, velocities, and other states.  
As seen in Fig.~\ref{fig:results2}, the robot accelerates to near the joint and end‑effector velocity limits (red dashed lines) to achieve a long throw, then decelerates.  
Peaks in velocity, acceleration, and torque occur near the throwing moment (blue dashed lines).

\subsection{Online Adaptation}
In this section, we demonstrate the online adaptability of our model in dynamically changing environments, enabled by its rapid planning speed.  
Consider a scenario where the robot is following an initially planned throwing trajectory for a given target box position, but the target position suddenly changes.  
We quickly replan a new throwing trajectory, allowing the robot to smoothly transition and track the updated trajectory without interruption.

\begin{figure}[!t]
    \centering
    \includegraphics[width=\linewidth]{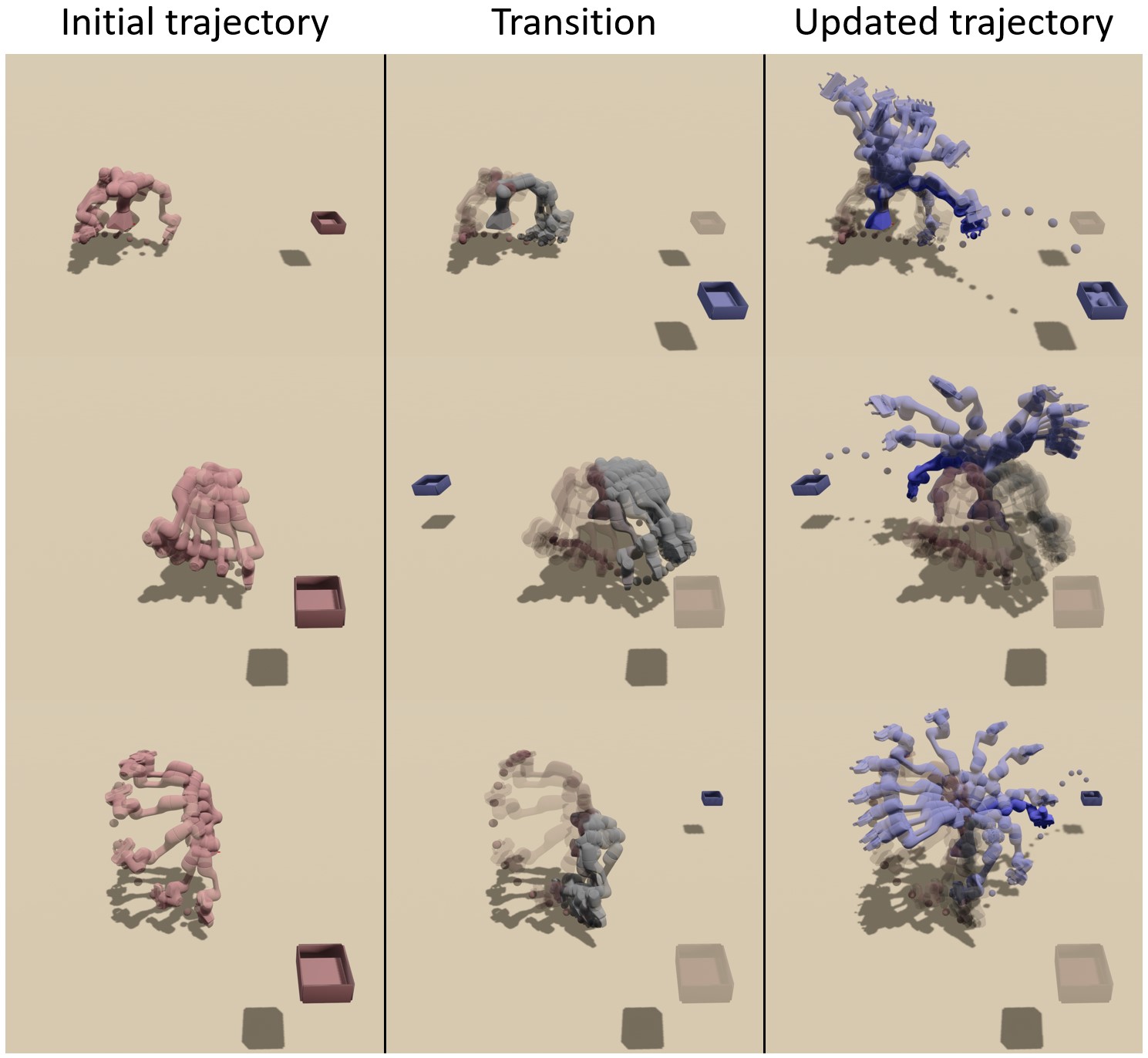}
    \caption{Online adaptation to changes in target box position: The robot follows an initial trajectory (red), detects a target change, replans an updated trajectory (blue), and executes a transition trajectory (gray) connecting its current phase to the new plan.}
    \label{fig:results3}
\end{figure}

In the replanning step, we sample 100 candidate trajectories $\{(q_i(t),\eta_i)\}_{i=1}^{100}$ conditioned on the new target.  
We then select the trajectory closest to the current configuration $q_c$ by solving
\[
(i^*,t^*) = \arg\min_{i,t}\|q_c-q_i(t)\| \quad \text{s.t. } t<\eta_i.
\]
Next, we construct a transition trajectory from the current phase $(q_c,\dot q_c)$ to $(q_{i^*}(t^*),\dot q_{i^*}(t^*))$ in the selected trajectory, using the following parametric model:
\begin{align}
\label{eq:mp2}
q(t) = &\, q_0 + (q_T-q_0)(3-2s)s^2 
+ \dot q_0 s - (2\dot q_0+\dot q_T)s^2 \nonumber\\
&+ (\dot q_0+\dot q_T)s^3 + s^2(s-1)^2\Phi(s)w,
\end{align}
in a manner analogous to~(\ref{eq:parmetric_model}).
This model enforces $(q_0,\dot q_0)$ at $t=0$ and $(q_T,\dot q_T)$ at $t=T$, and accommodates nonzero boundary velocities.

The transition trajectory must satisfy kinodynamic constraints, which reduces to finding a suitable $w$.  
Empirically, random Gaussian initialization of $w$ often yields at least one valid solution.  
The process is fast enough to run online, enabling the robot to adapt in real time.

Fig.~\ref{fig:results3} illustrates examples: red robots follow the initial plan until the target moves (after 1.8 s), gray robots execute a transition trajectory over $\sim$1 s, and blue robots follow the updated throwing trajectory, completing the throw within 3–5 s depending on target distance.

\section{Conclusion}

We have proposed Differentiable Motion Manifold Primitives (DMMP), which learns a manifold of continuous‑time, differentiable trajectories offline and enables fast kinodynamic motion planning through online search.  
We have trained DMMP through four steps: data collection, manifold learning, latent‑flow learning, and manifold optimization.  
Through dynamic throwing experiments with a 7‑DoF arm, we have demonstrated that DMMP can rapidly generate trajectories that satisfy both task objectives and kinodynamic constraints.

Our work can be further strengthened by improving the initial data‑collection step -- currently based on randomizing initial and final configurations -- as the diversity of collected trajectories directly determines the richness of the learned manifold and, consequently, the efficiency of online adaptation. Additionally, while this study focuses on planning, integrating tracking control and real‑world experiments would more comprehensively validate and demonstrate the effectiveness of the proposed approach.

\addtolength{\textheight}{-12cm}   






\bibliographystyle{IEEEtran}  
\bibliography{ref}

\end{document}